\documentclass[runningheads,a4paper]{llncs}

\usepackage{amsmath,amsfonts,amssymb}

\let\proof\relax
\let\endproof\relax

\usepackage{amsthm}
\usepackage{graphicx}
\usepackage{wrapfig}
\usepackage[percent]{overpic}
\usepackage{graphicx}
\usepackage{epsfig,epstopdf,url}

\usepackage{color}

\definecolor{pm}{rgb}{0.0,0.0,0.6}
\definecolor{jc}{rgb}{0.0,0.6,0.0}

\usepackage{balance}
\usepackage{enumerate}
\usepackage{url}

\newcommand{\keywords}[1]{\par\addvspace\baselineskip
\noindent\keywordname\enspace\ignorespaces#1}
\usepackage[colorlinks=true]{hyperref}

\newcommand{\ie}{\emph{i.e.}, }

\newtheorem{thm}{Theorem}

\usepackage{amsmath,amsfonts,amsthm}
\usepackage{amssymb}
\usepackage[commentsnumbered]{algorithm2e}
\usepackage{multirow}

\begin{document}
	

\title{A Compact Kernel Approximation for Efficient 3D Action Recognition}
\titlerunning{$\quad$}

%
%
\author{
Jacopo Cavazza$^{\mathbf{1},\mathbf{2}}$ \and Pietro Morerio$^{\mathbf{1}}$ \and Vittorio Murino$^{\mathbf{1},\mathbf{3}}$}
\authorrunning{$\quad$}

\institute{$^{\pmb 1}$Pattern Analysis \& Computer Vision (PAVIS) - Istituto Italiano di Tecnologia - \textit{Genova, Italy} \\
	$^{\pmb 2}$Electrical, Electronics and Telecommunication Engineering and Naval Architecture Department (DITEN) -- Universit\`{a} degli Studi di Genova --  \textit{Genova, Italy}\\
	$^{\pmb 3}$Computer Science Department -- Universit\`{a} di Verona --  \textit{Verona, Italy} \\
	{\tt\small firstname.lastname@iit.it}}

%
%

\maketitle

\begin{abstract}
3D action recognition was shown to benefit from a covariance representation of the input data (joint 3D positions). A kernel machine feed with such feature is an effective paradigm for 3D action recognition, yielding state-of-the-art results. Yet, the whole framework is affected by the well-known scalability issue. In fact, in general, the kernel function has to be evaluated for all pairs of instances inducing a Gram matrix whose complexity is quadratic in the number of samples. In this work we reduce such complexity to be linear by proposing a novel and explicit feature map to approximate the kernel function. This allows to train a linear classifier with an explicit feature encoding, which implicitly implements a Log-Euclidean machine in a scalable fashion. Not only we prove that the proposed approximation is unbiased, but also we work out an explicit strong bound for its variance, attesting a theoretical superiority of our approach with respect to existing ones. Experimentally, we verify that our representation provides a compact encoding and outperforms other approximation schemes on a number of publicly available benchmark datasets for 3D action recognition. 

\end{abstract}

\section{Introduction}\label{sec:intro}

Action recognition is a key research domain in video/image processing and computer vision, being nowadays ubiquitous in human-robot interaction, autonomous driving vehicles, elderly care and video-surveillance to name a few \cite{survey}. Yet, challenging difficulties arise due to visual ambiguities (illumination variations, texture of clothing, general background noise, view heterogeneity, occlusions). As an effective countermeasure, joint-based skeletal representations (extracted from depth images) are a viable solution.

Combined with a skeletal representation, the symmetric and positive definite (SPD) covariance operator scores a sound performance in 3D action recognition \cite{Wang:ICCV15,Harandi:CVPR14,Cavazza:ICPR16}. Indeed, while properly modeling the skeletal dynamics with a second order statistic, the covariance operator is also naturally able to handle different temporal duration of action instances. This avoids slow pre-processing stages such as time warping or interpolation \cite{Vemulapalli:CVPR14}. In addition, the superiority of such representation can be attested by achieving state-of-the-art performance by means of a relative simple classification pipeline \cite{Wang:ICCV15,Cavazza:ICPR16} where, basically\footnote{For the sake of precision, let us notice that \cite{Wang:ICCV15} take advantage of multiple kernel learning in combining several low-level representations and \cite{Cavazza:ICPR16} replaces the classical covariance operator with a kernelization.}, a non-linear Support Vector Machine (SVM) is trained using the Log-Euclidean kernel
\begin{equation}\label{eq:KlE}
K_{\ell E}(\mathbf{X},\mathbf{Y}) = \exp \left( - \dfrac{1}{2 \sigma^2} \| \log \mathbf{X} - \log \mathbf{Y} \|_F^2\right)
\end{equation}
to compare covariance operators $\mathbf{X}$, $\mathbf{Y}$. In \eqref{eq:KlE}, for any SPD matrix $\mathbf{X}$, we define 
\begin{equation}\label{eq:log}
\log \mathbf{X} = \mathbf{U} {\rm diag} (\log \lambda_1, \dots, \log \lambda_d) \mathbf{U}^\top,
\end{equation}
being $\mathbf{U}$ the matrix of eigenvectors which diagonalizes $\mathbf{X}$ in terms of the eigenvalues $\lambda_1 \geq \dots \geq \lambda_d > 0$. Very intuitively, for any fixed bandwidth $\sigma > 0$, $K_{\ell E}(\mathbf{X},\mathbf{Y})$ is actually computing a radial basis Gaussian function by comparing the covariance operators $\mathbf{X}$ and $\mathbf{Y}$ by means of the Frobenius norm $\| \cdot \|_F$ (after $\mathbf{X},\mathbf{Y}$ have been log-projected). Computationally, the latter stage is not problematic (see Section \ref{sez:RGW}) and can be performed for each covariance operator \emph{before} computing the kernel. In addition to its formal properties in Riemannian geometry, this makes \eqref{eq:KlE} widely used in practice \cite{Harandi:CVPR14,Wang:ICCV15,Cavazza:ICPR16}.

However, the modern big data regime mines the applicability of such a kernel function. Indeed, since \eqref{eq:KlE} has to be computed for \emph{every pair} of instances in the dataset, the so produced Gram matrix has prohibitive size. So its storage becomes time- and memory-expensive and the related computations (required to train the model) are simply unfeasible.

The latter inconvenient can be solved as follows. According to the well established kernel theory \cite{Bishop}, the Kernel \eqref{eq:KlE} induces an infinite-dimension feature map $\varphi$, meaning that $K_{\ell E}(\mathbf{X},\mathbf{Y}) = \langle \varphi(\mathbf{X}), \varphi(\mathbf{Y}) \rangle$. However, if we are able to obtain an explicit feature map $\Phi$ such that $K_{\ell E}(\mathbf{X},\mathbf{Y}) \approx \langle \Phi(\mathbf{X}), \Phi(\mathbf{Y}) \rangle$, we can directly compute a finite-dimensional feature representation $\Phi(\mathbf{X})$ for each action instance separately. Then, with a compact $\Phi$, we can train a linear SVM instead of its kernelized version. This is totally feasible and quite efficient even in the big data regime \cite{liblinear}. Therefore, the whole pipeline will actually provide a scalable implementation of a Log-Euclidean SVM, whose cost is reduced from quadratic to linear.
 
In our work we specifically tackle the aforementioned issue through the following main contributions.

\begin{enumerate}
	\item We propose a novel compact and explicit feature map to approximate the Log-Euclidean kernel within a probabilistic framework. 
	\item We provide a rigorous mathematical formulation, proving that the proposed approximation has null bias and bounded variance.
	\item We compare the proposed feature map approximation against alternative approximation schemes, showing the formal superiority of our framework.
	\item We experimentally evaluate our method against the very same approximation schemes over six 3D action recognition datasets, confirming with practice our theoretical findings.
\end{enumerate}

The rest of the paper is outlined as follows. In Section \ref{sez:RW} we review the most relevant related literature. Section \ref{sez:RGW} proposes the novel approximation and discusses its foundation. We compare it with alternative paradigms in Section \ref{sez:practice}. Section \ref{sec:concl} draws  conclusions and the Appendix \ref{sez:app} reports all proofs of our theoretical results.
 
 \section{Related work}\label{sez:RW}
 
 In this Section, we summarize the most relevant works in both covariance-based 3D action recognition and kernels' approximations. 
 
Originally envisaged for image classification and detection tasks, the covariance operator has experienced a growing interest for action recognition, experiencing many different research trends: \cite{Harandi:CVPR14} extends it to the infinite dimensional case, while  \cite{egizi} hierarchically combines it in a temporal pyramid; \cite{Wang:ICCV15,ECCV16} investigate the conceptual analogy with trial-specific kernel matrices and \cite{Cavazza:ICPR16} further proposes a new kernelization as to model arbitrary, non-linear relationships conveyed by the raw data. 
However, those kernel methods usually do not scale up easily to big datasets due to demanding storage and computational costs. As a solution, the exact kernel representation can be replaced by an approximated, more efficient version. In the literature, this is done according to the following mainstream approaches. 
 \vspace{-.2 cm}
 \begin{enumerate}[$(i)$]	
 \itemsep0em
 \item The kernel Gram matrix is replaced with a surrogate low-rank version, in order to alleviate both memory and computational costs. Within these methods, \cite{Bach:ICML05} applied Cholesky decomposition and \cite{Zhang:ICML08} exploited Nystr\"{o}m approximation. 
 \item Instead of the exact kernel function $k$, an explicit feature map $\Phi$ is computed, so that the induced linear kernel $\langle \Phi(\mathbf{x}),\Phi(\mathbf{y}) \rangle$ approximates $k(\mathbf{x},\mathbf{y})$. Our work belong to this class of methods. 
 \end{enumerate}  \vspace{-.2 cm}
 In this context, Rahimi \& Recht \cite{RR:NIPS07} exploited the formalism of the Fourier Transform to approximate shift invariant kernels $k(\mathbf{x},\mathbf{y}) = k(\mathbf{x}-\mathbf{y})$ through an expansion of trigonometric functions. Leveraging on a similar idea, Le et al. \cite{Fastfood} sped up the computation by exploiting the Walsh-Hadamard transform, downgrading the running cost of \cite{RR:NIPS07} from linear to log-linear with respect to the data dimension. Recently, Kar \& Karnick \cite{KK:AISTATS12} proposed an approximated feature maps for dot product kernels $k(\mathbf{x},\mathbf{y}) = k(\langle \mathbf{x},\mathbf{y} \rangle)$ by directly exploiting the MacLaurin expansion of the kernel function.
 
Instead of considering a generic class of kernels, our work specifically focuses on the log-Euclidean one, approximating it through a novel unbiased estimator which ensures a explicit bound for variance (as only provided by \cite{Fastfood}) and resulting in a superior classification performance with respect to \cite{RR:NIPS07,Fastfood,KK:AISTATS12}.

\section{The proposed approximated feature map}\label{sez:RGW}

In this Section, we present the main theoretical contribution of this work, namely i) a random, explicit feature map $\Phi$ such that $\langle \Phi(\mathbf{X}), \Phi(\mathbf{Y}) \rangle \approx K_{\ell E}(\mathbf{X},\mathbf{Y})$, ii) the proof of its unbiasedness and iii) a strong theoretical bound on its variance.

{\bf Construction of the approximated feature map.} In order to construct a $\nu$ dimensional feature map $\mathbf{X} \mapsto \Phi(\mathbf{X}) = [\Phi_1(\mathbf{X}),\dots,\Phi_\nu(\mathbf{X})] \in \mathbb{R}^{\nu}$, for any $d \times d$ SPD matrix $\mathbf{X}$, fix a probability distribution $\rho$ supported over $\mathbb{N}$. Precisely, each component $\Phi_1, \dots, \Phi_\nu$ of our $\nu$-dimensional feature map $\Phi$ is independently computed according to the following algorithm. 
\vspace{-.7 cm}
\begin{algorithm}[h!]
	\ForEach{$j = 1,\dots,\nu$}{
		\nl Sample $n$ according to $\rho$\\
		\nl Sample the $d^n \times d^n$ matrix $\mathbf{W}$ of independent Gaussian distributed weights with zero mean and $\sigma^2/\sqrt{n}$ variance \\
		\nl Compute $\log(\mathbf{X})^{\otimes n} = \log \mathbf{X} \otimes \dots \otimes \log\mathbf{X}$, $n$ times. \\
		\nl Assign \vspace{-.2 cm}	\begin{equation}\label{eq:varphi}
		\Phi_j(\mathbf{X}) = \dfrac{1}{\sigma^{2n}}\sqrt{ \dfrac{\exp(-\sigma^{-2})}{\nu \rho(n) n!} } {\rm tr}( \mathbf{W}^\top \log(\mathbf{X})^{\otimes n}). \vspace{-.2 cm}
		\end{equation}}\vspace{-.7 cm}
\end{algorithm}

The genesis of \eqref{eq:varphi} can be explained by inspecting the feature map $\varphi$ associated to the kernel $K(x,y)=\exp\left(-\frac{1}{2\sigma^2}|x-y|^2\right)$, where $x,y \in \mathbb{R}$ for simplicity. It results 
$\varphi(x) \propto \left[ 1, \sqrt{\frac{1}{1!\sigma^2}}x,\sqrt{\frac{1}{2!\sigma^4}}x^2,\sqrt{\frac{1}{3!\sigma^6}}x^3,\dots\right].$
Intuitively, we can say that \eqref{eq:varphi} approximates the infinite dimensional $\varphi(x)$ by randomly selecting one of its components: this is the role played by $n \sim \rho$. In addition, we introduce the $\log$ mapping and replace the exponentiation with a Kronecker product. As a consequence, the random weights $\mathbf{W}$ ensure that $\Phi(\mathbf{X})$ achieves a sound approximation of \eqref{eq:KlE}, in terms of unbiasedness and rapidly decreasing variance. \\
In the rest of the Section we discuss the theoretical foundation of our analysis, where all proofs have been moved to Appendix \ref{sez:app} for convenience.

{\bf Unbiased estimation.} In order for a statistical estimator to be reliable, we need it to be at least \textit{unbiased}, \ie its expected value must be equal to the exact function it is approximating. The unbiasedness of the feature map $\Phi$ of eq. \eqref{eq:varphi} for the Log-Euclidean kernel \eqref{eq:KlE} is proved by the following result.

\begin{thm}\label{thm:Phi}
	Let $\rho$ be a generic probability distribution over $\mathbb{N}$ and consider $\mathbf{X}$ and $\mathbf{Y}$, two generic SPD matrices such that $\| \log \mathbf{X} \|_F = \| \log \mathbf{Y} \|_{F} = 1.$ Then, $\langle \Phi(\mathbf{X}), \Phi(\mathbf{Y}) \rangle$ is an unbiased estimator of \eqref{eq:KlE}. That is \begin{equation}\label{eq:meanned}
	\mathbb{E}[\langle \Phi(\mathbf{X}), \Phi(\mathbf{Y}) \rangle] = K_{\ell E} (\mathbf{X}, \mathbf{Y}),
	\end{equation} 
	where the expectation is computed over $n$ and $\mathbf{W}$ which define $\Phi_j(\mathbf{X})$ as in \eqref{eq:varphi}.
\end{thm}

Once averaging upon all possible realizations of $n$ sampled from $\rho$ and the Gaussian weights $\mathbf{W}$, Theorem \ref{thm:Phi} guarantees that the linear kernel $\langle \Phi(\mathbf{X}), \Phi(\mathbf{Y}) \rangle$ induced by $\Phi$ is equal to $K_{\ell E}(\mathbf{X},\mathbf{Y})$. This formalizes the unbiasedness of our approximation.

\textit{On the assumption} $\| \log \mathbf{X} \|_F = \| \log \mathbf{Y} \|_{F} = 1$. Under a practical point of view, this assumption may seem unfavorable, but this is not the case. The reason is provided by equation \eqref{eq:log}, which is very convenient to compute the logarithm of a SPD matrix. Since in \eqref{eq:varphi}, $\Phi(\mathbf{X})$ is explicitly dependent on $\log \mathbf{X}$, we can simply use \eqref{eq:log} and then divide each entry of the obtained matrix by $\| \log \mathbf{X} \|_F$. This is a non-restrictive strategy to satisfy our assumption and actually analogous to require input vectors to have unitary norm, which is very common in machine learning \cite{Bishop}.

{\bf Low-variance.} One can note that, in Theorem \ref{thm:Phi}, even by choosing $\nu = 1$ (a scalar feature map), $\Phi(\mathbf{X}) = [\Phi_1(\mathbf{X})] \in \mathbb{R}$ is unbiased for \eqref{eq:KlE}. However, since $\Phi$ is an approximated finite version of the exact infinite feature map associated to \eqref{eq:KlE}, one would expect the quality of the approximation to be very bad in the scalar case, and to improve as $\nu$ grows larger. This is indeed true, as proved by the following statement.

\begin{thm}\label{th:var}
	The variance of $\langle \Phi(\mathbf{X}), \Phi(\mathbf{Y}) \rangle$ as estimator of \eqref{eq:KlE} can be explicitly bounded. Precisely, 
	\begin{equation}\label{eq:var}
	\mathbb{V}_{n,\mathbf{W}}(K_\Phi(\mathbf{X},\mathbf{Y})) \leq \dfrac{\mathcal{C}_\rho}{\nu^3} \exp\left(\dfrac{3 - 2\sigma^2}{\sigma^4}\right),
	\end{equation}
	where $\| \log \mathbf{X} \|_F = \| \log \mathbf{Y} \|_F = 1$ and the variance is computed over all possible realizations of $n \sim \rho$ and $\mathbf{W}$, the latter being element-wise sampled from a $\mathcal{N}(0,\sigma^2/\sqrt{n})$ distribution. Also, $\mathcal{C}_\rho \stackrel{\rm def}{=} \sum_{n = 0}^{\infty} \frac{1}{\rho(n) n!}$, the series being convergent.
\end{thm}

Let us discuss the bound on the variance provided by Theorem \ref{th:var}. Since the bandwidth $\sigma$ of the kernel function \eqref{eq:KlE} we want to approximate is fixed, the term $\exp\hspace{-.5 mm} \left(\hspace{-.5 mm} \frac{3 \hspace{-.2 mm} - \hspace{-.2 mm}  2\sigma^2}{\sigma^4}\hspace{-.5 mm} \right)$ can be left out from our analysis. The bound in \eqref{eq:var} is linear in $\mathcal{C}_\rho$ and inversely cubic in $\nu$. When $\nu$ grows, the increased dimensionality of our feature map $\Phi$ makes the variance rapidly vanishing, ensuring that the \textit{approximated kernel} $K_\Phi(\mathbf{X}, \hspace{-.5 mm} \mathbf{Y})= \langle \Phi(\mathbf{X}), \Phi(\mathbf{Y}) \rangle $ converges to the target one, that is $K_{\ell E}$. Such trend may be damaged by big values of $\mathcal{C}_\rho.$ Since the latter depends on the distribution $\rho$, let us fix it to be the geometric distribution $\mathcal{G}(\theta)$ with parameter $0 \leq \theta < 1$. This yields

\begin{equation}\label{eq:C}
\mathcal{C}_\rho \propto \sum_{n = 0}^\infty \dfrac{1}{(1 - \theta)^n \cdot n!} = \exp\left( \dfrac{1}{1 - \theta} \right).
\end{equation}

There is a tradeoff between a low variance (\ie $\mathcal{C}_\rho$ small) and a reduced computational cost for $\Phi$ (\ie $n$ small). Indeed, choosing $\theta \approx 1$ makes $\mathcal{C}_\rho$ big in \eqref{eq:C}. In this case, the integer $n$ sampled from $\rho = \mathcal{G}(\theta)$ is small with great probability: this leads to a reduced number of Kronecker products to be computed in $\log(\mathbf{X})^{\otimes n}$. Conversely, when $\theta \approx 0$, despite $n$ and the related computational cost of $\log(\mathbf{X})^{\otimes n}$ are likely to grow, $\mathcal{C}_\rho$ is small, ensuring a low variance for the estimator.

As a final theoretical result, Theorems \ref{thm:Phi} and \ref{th:var} immediately yield that
\begin{equation}\label{eq:abs}
\hspace{-.4 cm}\mathbb{P}\hspace{-.5 mm}\left[ \left| \hspace{-.5 mm} K_\Phi(\mathbf{X}, \hspace{-.5 mm} \mathbf{Y}) \hspace{-.5 mm} - \hspace{-.5 mm} K_{\ell E}(\mathbf{X}, \hspace{-.5 mm} \mathbf{Y}) \hspace{-.5 mm} \right| \hspace{-.5 mm} \geq  \hspace{-.5 mm}\epsilon \right] \hspace{-.5 mm} \leq \hspace{-.5 mm}  \frac{\mathcal{C}_\rho}{\nu^3 \epsilon^2}\hspace{-.5 mm} \exp \hspace{-.5 mm}\left(\hspace{-.7 mm}\frac{3\hspace{-.5 mm} -\hspace{-.5 mm} 2\sigma^2}{\sigma^4}\hspace{-.7 mm}\right)\hspace{-.5mm}
\end{equation}
for every pairs of unitary Frobenius normed SPD matrices $\mathbf{X},\mathbf{Y}$ and $\epsilon > 0$, as a straightforward implication of the Chebyshev inequality. This ensures that $K_\Phi$ differs in module from $K_{\ell E}$ by more than $\epsilon$ with a (low) probability $\mathbb{P}$, which is inversely cubic and quadratic in $\nu$ and $\epsilon$, respectively.

{\bf \textit{Final remarks}.} To sum up, we have presented a constructive algorithm to compute a  $\nu$-dimensional feature map $\Phi$ whose induced linear kernel is an unbiased estimator of the log-Euclidean one. Additionally, we ensure an explicit bound on the variance which rapidly vanishes as $\nu$ grows (inversely cubic decrease). This implies that $\langle \Phi(\mathbf{X}), \Phi(\mathbf{Y}) \rangle$ and $K_{\ell E}(\mathbf{X},\mathbf{Y})$ are equal with very high probability, even at low $\nu$ values. This implements a Log-Euclidean kernel in a scalable manner, downgrading the quadratic cost of computing $K_{\ell E}(\mathbf{X},\mathbf{Y})$ for every $\mathbf{X},\mathbf{Y}$ into the linear cost of evaluating the feature map $\Phi(\mathbf{X})$ as in \eqref{eq:varphi} for every $\mathbf{X}$. Practically, this achieve a linear implementation of the log-Euclidean SVM.



\section{Results}\label{sez:practice}
In this Section, we compare our proposed approximated feature map versus the alternative ones by Rahimi \& Recht \cite{RR:NIPS07}, Kar \& Karnick \cite{KK:AISTATS12} and Le et al. \cite{Fastfood} (see Section \ref{sez:RW}). 

{\bf Theoretical Comparison.} Let us notice that all approaches \cite{RR:NIPS07,KK:AISTATS12,Fastfood} are applicable also to the log-Euclidean kernel \eqref{eq:KlE}. Indeed, \cite{RR:NIPS07,Fastfood} includes our case of study since $K_{\ell E}(\mathbf{X},\mathbf{Y}) = k(\log \mathbf{X} - \log \mathbf{Y})$ is logarithmic shift invariant. At the same time, thanks to the assumption $\| \log \mathbf{X} \|_F = \| \log \mathbf{Y} \|_F = 1$ as in Theorem \ref{thm:Phi}, we obtain $K_{\ell E}(\mathbf{X},\mathbf{Y}) = k(\langle \log \mathbf{X}, \log \mathbf{Y} \rangle)$ (see \eqref{eq:huhu} in Appendix \ref{sez:app}), thus satisfying the hypothesis of Kar \& Karnick \cite{KK:AISTATS12}.

As we proved in Theorem \ref{thm:Phi}, all works \cite{RR:NIPS07,KK:AISTATS12,Fastfood} can also guarantee an unbiased estimation for the exact kernel function.

Actually, what makes our approach superior is the explicit bound on the variance (see Table \ref{tab:b}). Indeed, \cite{RR:NIPS07,KK:AISTATS12} are totally lacking in this respect. Moreover, despite an analogous bound is provided in \cite[Theorem 4]{Fastfood}, it only ensures a $O(1/\nu)$ convergence rate for the variance with respect to the feature dimensionality $\nu$. Differently, we can guarantee a $O(1/{\nu^3})$ trend. This implies that, \textit{we achieve a better approximation of the kernel with a lower dimensional feature representation}, which ease the training of the linear SVM \cite{liblinear}.

\begin{table}[t!]
	\centering
	\begin{tabular}{|c|c|c|c|}
		\hline
		\textit{proposed} & Rahimi \& Recht \cite{RR:NIPS07} & Kar \& Karninck \cite{KK:AISTATS12} & Le et al. \cite{Fastfood} \\ \hline
		$O(1/\nu^3)$ & \texttt{missing} & \texttt{missing} & $O(1/\nu)$ \\\hline
	\end{tabular}
	\caption{Theoretical comparison between explicit bounds on variance between the proposed approximation and \cite{RR:NIPS07,KK:AISTATS12,Fastfood}: the quicker the decrease, the better the bound. Here, $\nu$ denotes the dimensionality of the approximated feature vector.}
	\label{tab:b}
\end{table}

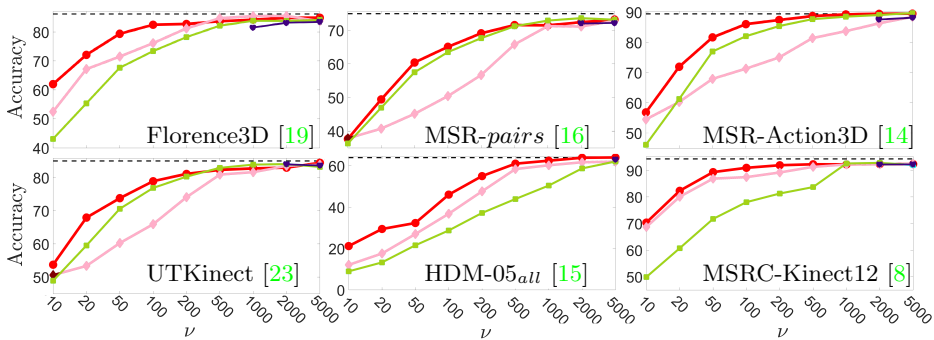
\begin{figure}[t!]
	\begin{overpic}[width=\textwidth]{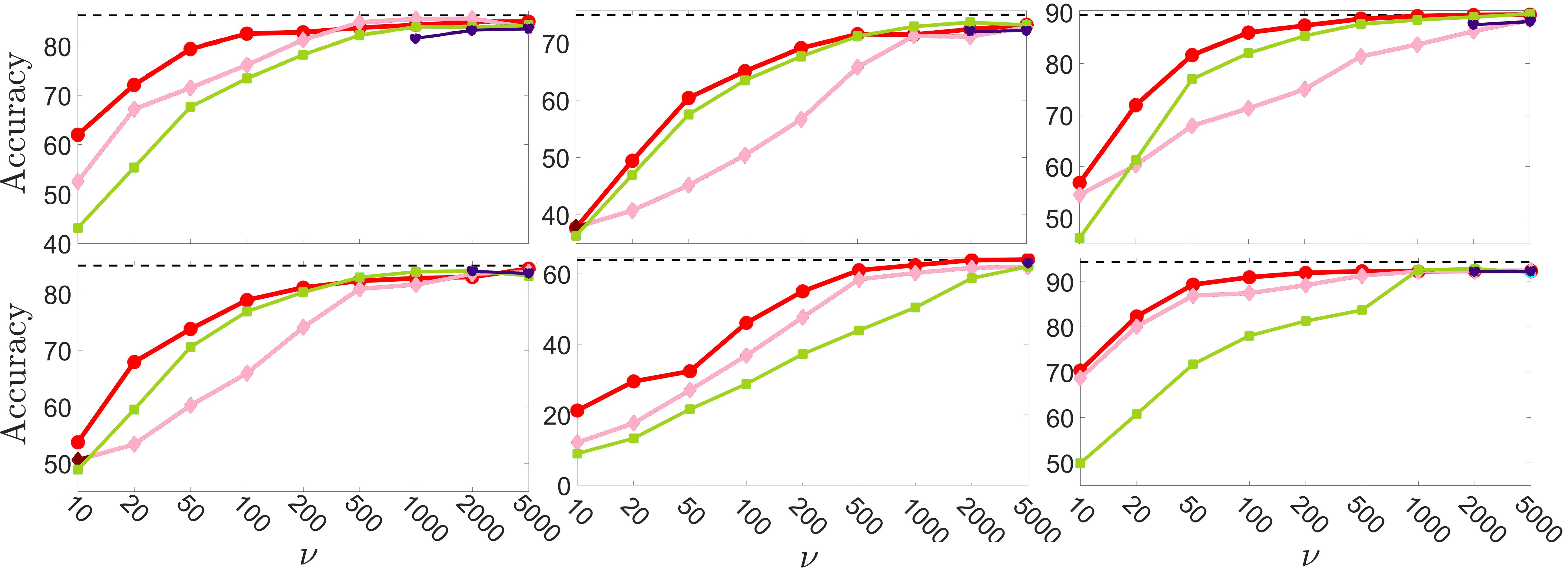}
		\put (15,22) {Florence3D \cite{Florence3D}}
		\put (45,22) {MSR-$pairs$ \cite{MSRPairs}}
		\put (75,22) {MSR-Action3D \cite{Action3D}}
		\put (15,7) {UTKinect \cite{UTKinect}}
		\put (45,7) {HDM-05$_{all}$ \cite{HDM-05}}
		\put (75,7) {MSRC-Kinect12 \cite{MSRC}}
	\end{overpic}
	\caption{Experimental comparison of our approximation (red curves) against the schemes ofr Rahimi \& Recht \cite{RR:NIPS07} (pink curves), Kar \& Karnick \cite{KK:AISTATS12} (green curves) and Le et al. \cite{Fastfood} (blue curves). Best viewed in colors.}
	\label{fig:1}
\end{figure}

{\bf Experimental Comparison.} We reported here the experimental comparison on 3D action recognition between our proposed approximation and the paradigms of \cite{RR:NIPS07,KK:AISTATS12,Fastfood}. 

\textit{Datasets.} For the experiments, we considered UTKinect \cite{UTKinect}, Florence3D \cite{Florence3D}, MSR-Action-Pairs (MSR-$pairs$) \cite{MSRPairs}, MSR-Action3D \cite{Action3D}, \cite{G3D}, HDM-05 \cite{HDM-05} and MSRC-Kinect12 \cite{MSRC} datasets.\\
For each, we follow the usual training and testing splits proposed in the literature. For Florence3D and UTKinect, we use the protocols of \cite{Vemulapalli:CVPR14}. For MSR-Action3D, we adopte the splits originally proposed by \cite{Action3D}. On MSRC-Kinect12, once highly corrupted action instances are removed as in \cite{egizi}, training is performed on odd-index subject, while testing on the even-index ones. On HDM-05, the training exploited all instances of ``\texttt{bd}'' and ``\texttt{mm}'' subjects, being  ``\texttt{bk}'', ``\texttt{dg}'' and ``\texttt{tr}'' left out for testing \cite{Wang:ICCV15}, using the 65 action classes protocol of \cite{bellolui}. 

\textit{Data preprocessing.} As a common pre-processing step, we normalize the data by computing the relative displacements of all joints $x-y-z$ coordinates and the ones of the hip (central) joint, for each timestamp. 

\textit{Results.} Figure \ref{fig:1} reports the quantitative performance while varying $\nu$ in the range 10, 20, 50, 100, 200, 500, 1000, 2000, 5000. When comparing with \cite{Fastfood}, since the data input size must be a multiple of a power of 2, we zero-padded our vectorized representation to match 4096 and (whenever possible) 2048 and 1024 input dimensionality. These cases are then compared with the results related to $\nu =$ 5000, 2000, 1000 for RGW and \cite{RR:NIPS07,KK:AISTATS12}, respectively. Since all approaches have a random component, we performed ten repetitions for each method and dimensionality setup, averaging the scored classification performances obtained through a linear SVM with $C = 10$. We employ the publicly available codes for \cite{RR:NIPS07,KK:AISTATS12,Fastfood}. Finally, we also report the classification performance with the exact method obtained by feeding an SVM with the log-Euclidean kernel whose bandwidth $\sigma$ is chosen via cross validation.

{\bf \textit{Discussion.}} For large $\nu$ values, all methods are able to reproduce the performance of the log-Euclidean kernel (black dotted line). Still, in almost all the cases, our approximation is able to outperform the competitors: for instance, we gapped Rahimi and Recht on both MSR-Pairs and MSR-Action3D, while Kar \& Karnick scored a much lower performance on HDM-05 and Florence3D. If comparing to Le et al., the performance is actually closer, but this happens for all the methods which are able to cope the performance of the Log-Euclidean kernel with $\nu \geq 2000,5000$. Precisely, the true superiority of our approach is evident in the case of a small $\nu$ value ($\nu =$ 10, 20, 50). Indeed, our approximation always provides a much rapid growing accuracy (MSR-Action3D, Florence3D and UTKinect), with only a few cases where the gap is thinner (Kar \& Karnick \cite{KK:AISTATS12} on MSR-$pairs$ and Rahimi \& Recth \cite{RR:NIPS07} on MSRC-Kinect 12). Therefore, our approach ensures a more descriptive and compact representation, providing a superior classification performance.

\section{Conclusions}\label{sec:concl}

In this work we propose a novel scalable implementation of a Log-Euclidean SVM to perform proficient classification of SPD (covariance) matrices. We achieve a linear complexity by providing an explicit random feature map whose induced linear kernel is an unbiased estimator of the exact kernel function.

Our approach proved to be more effective than alternative approximations \cite{RR:NIPS07,KK:AISTATS12,Fastfood}, both theoretically and experimentally. Theoretically, we achieve an explicit bound on the variance on the estimator (such result is totally absent in \cite{RR:NIPS07,KK:AISTATS12}), which is decreasing with inversely cubic pace versus the inverse linear of \cite{Fastfood}. Experimentally, through a broad evaluation, we assess the superiority of our representation which is able to provide a superior classification performance at a lower dimensionality.

\appendix

\section{Proofs of all theoretical results}\label{sez:app}

In this Appendix we report the formal proofs for both the unbiased approximation (Theorem \ref{thm:Phi}) and the related rapidly decreasing variance (Theorem \ref{th:var}).

\proof[Proof of Theorem \ref{thm:Phi}] Use the definition of \eqref{eq:varphi} and the linearity of the expectation. We get that $\mathbb{E}_{n,\mathbf{W}}\left[ \langle \Phi(\mathbf{X}), \Phi(\mathbf{Y}) \rangle \right]$ equals to
\begin{equation}\label{eq:hacca}
\mathbb{E}_{n}\left[ \dfrac{1}{\sigma^{4n}} \dfrac{\exp(-\sigma^{-2})}{\rho(n) n!} \mathbb{E}_{\mathbf{W}} \left[ {\rm tr}\left( \mathbf{W}^\top \log(\mathbf{X})^{\otimes n} \right) {\rm tr} \left( \mathbf{W}^\top \log(\mathbf{Y})^{\otimes n}\right) \right]\right],
\end{equation}
by simply noticing that the dependence with respect to $\mathbf{W}$ involves the terms inside the trace operators only. Let us focus on the term ${\rm tr}\left( \mathbf{W}^\top \log(\mathbf{X})^{\otimes n} \right)$. We can expand 
\begin{equation}\label{eq:pippo}
{\rm tr}\left( \mathbf{W}^\top \log(\mathbf{X})^{\otimes n} \right) = \sum_{i_1,\dots,i_{2n} = 1}^d w_{i_1,\dots,i_{2n}} \log(\mathbf{X})_{i_1,i_2} \cdots \log(\mathbf{X})_{i_{2n -1},i_{2n}}
\end{equation}
by using the definition of $\log(\mathbf{X})^{\otimes n}$ and the properties of the trace operator. In equation \eqref{eq:pippo}, we replace the random coefficient $w_{i_1,\dots,i_{2n}}$ with $u^{(1)}_{i_1,i_2},\dots,u^{(n)}_{i_{2n - 1},i_{2n}}$ independent and identically distributed (i.i.d.) according to a $\mathcal{N}(0,\sigma^2)$ distribution. We can notice that \eqref{eq:pippo} can be rewritten as
\begin{equation}\label{eq:luca}
{\rm tr}\left( \mathbf{W}^\top \log(\mathbf{X})^{\otimes n} \right) = \prod_{\alpha = 1}^{n} \sum_{i,j = 1}^d u^{(\alpha)}_{i,j} \log(\mathbf{X})_{ij}.
\end{equation}
Making use of \eqref{eq:luca} in \eqref{eq:hacca}, we can rewrite $\mathbb{E}_{n,\mathbf{W}}\left[ K_\Phi(\mathbf{X},\mathbf{Y})\right]$ as
\begingroup\makeatletter\def\f@size{9}\check@mathfonts
\begin{align}
 \mathbb{E}_{n}\left[ \dfrac{1}{\sigma^{4n}} \dfrac{\exp(-\sigma^{-2})}{\rho(n) n!} \mathbb{E}_{\mathbf{W}} \left[ \left( \sum_{i,j = 1}^d u^{(1)}_{i,j} \log(\mathbf{X})_{ij} \right)\left( \sum_{h,k = 1}^d u^{(1)}_{h,k} \log(\mathbf{Y})_{hk} \right) \right]^n \right] \label{eq:hacca2}
\end{align}\endgroup
by also considering the independence of $u^{(\alpha)}_{i,j}$ are independent. By furthermore using the fact that $\mathbb{E}_{\mathbf{W}}\left[ u^{(1)}_{i,j}u^{(1)}_{h,k} \right] = 0$ if $i \neq h$ and $j \neq k$ and the formula for the variance of a Gaussian distribution, we get
\begin{align}
\mathbb{E}_{n,\mathbf{W}}\left[ K_\Phi(\mathbf{X},\mathbf{Y})\right] = \mathbb{E}_{n}\left[ \dfrac{1}{\sigma^{4n}} \dfrac{\exp(-\sigma^{-2})}{\rho(n) n!} \sigma^{2n} \left( \langle \log(\mathbf{X}), \log(\mathbf{Y}) \rangle_F \right)^{n} \right], \label{eq:hacca3}
\end{align}
by introducing the Frobenius inner product $\langle \mathbf{A}, \mathbf{B} \rangle_F = \sum_{i,j = 1}^d \mathbf{A}_{ij} \mathbf{B}_{ij}$ between matrices $\mathbf{A}$ and $\mathbf{B}$. By expanding the expectation over $\rho$, \eqref{eq:hacca3} becomes
\begin{align}
\mathbb{E}_{n,\mathbf{W}}\left[ K_\Phi(\mathbf{X},\mathbf{Y})\right] &= \sum_{n = 0}^\infty \rho(n) \dfrac{1}{\sigma^{2n}} \dfrac{\exp(-\sigma^{-2})}{\rho(n)n!} (\langle \log(\mathbf{X}), \log(\mathbf{Y}) \rangle_F)^n \nonumber \\ &= \exp\left(-\dfrac{1}{\sigma^{2}}\right) \sum_{n = 0}^\infty \left(\dfrac{\langle \log(\mathbf{X}), \log(\mathbf{Y}) \rangle_F}{\sigma^{2}}\right)^n \dfrac{1}{n!}. \label{eq:huhu}
\end{align}  
The thesis easily comes from \eqref{eq:huhu} by using the Taylor expansion for the exponential function and the assumption $\|\log(\mathbf{X})\|_F = \|\log(\mathbf{Y})\|_F = 1$.
\endproof

\hrulefill

\proof[Proof of Theorem \ref{th:var}] Due to the independence of the components in $\Phi$, by definition of inner product we get $
\mathbb{V}_{n,\mathbf{W}} \left[ \langle \Phi(\mathbf{X}),\Phi(\mathbf{Y}) \rangle \right] = \nu \mathbb{V}_{n,\mathbf{W}} \left[  \Phi_1(\mathbf{X}) \Phi_1(\mathbf{Y})  \right]$. But then $\mathbb{V}_{n,\mathbf{W}} \left[ \langle \Phi(\mathbf{X}),\Phi(\mathbf{Y}) \rangle \right] \leq \nu \mathbb{E}_{n,\mathbf{W}} \left[  \Phi_1(\mathbf{X})^2 \Phi_1(\mathbf{Y})^2  \right]$ by definition of variance. Taking advantage of \eqref{eq:varphi}, yields to the equality between $\mathbb{V}_{n,\mathbf{W}} \left[ K_\Phi(\mathbf{X},\mathbf{Y}) \right]$ and
\begingroup\makeatletter\def\f@size{9}\check@mathfonts
\begin{equation}\label{eq:koala}
\dfrac{1}{\nu^3} \mathbb{E}_{n,\mathbf{U}} \left[ \dfrac{1}{\sigma^{8n}} \dfrac{\exp(-2\sigma^{-2})}{(\rho(n) n!)^2} \prod_{\alpha = 1}^{n} \left( \sum_{i,j = 1}^d u^{(\alpha)}_{i,j} \log(\mathbf{X})_{ij} \right)^{\hspace{-.1 cm} 2} \hspace{-.1 cm} \left( \sum_{h,k = 1}^d u^{(\alpha)}_{h,k} \log(\mathbf{Y})_{hk} \right)^{\hspace{-.1 cm} 2} \right], 
\end{equation}\endgroup
where $u^{(1)}_{i_1,i_2},\dots,u^{(n)}_{i_{2n - 1},i_{2n}}$ are i.i.d. from $\mathcal{N}(0,\sigma^2)$ distribution used to re-parametrize the original weights $\mathbf{W}$. Exploit the independence of $u^{(\alpha)}_{ij}$ to rewrite \eqref{eq:koala} as
\begingroup\makeatletter\def\f@size{9}\check@mathfonts
\begin{align}
\dfrac{1}{\nu^3} \mathbb{E}_{n} \left[ \dfrac{1}{\sigma^{8n}} \dfrac{\exp(-2\sigma^{-2})}{(\rho(n) n!)^2} \mathbb{E}_{\mathbf{U}} \left[ \left( \sum_{i,j = 1}^d u^{(1)}_{i,j} \log(\mathbf{X})_{ij} \right)^{\hspace{-5pt }2}\left( \sum_{h,k = 1}^d u^{(1)}_{h,k} \log(\mathbf{Y})_{hk} \right)^{\hspace{-5pt} 2} \right]^{n}\right]. \label{eq:cactus}
\end{align}\endgroup
By exploiting the zero correlation of the weights in $\mathbf{U}$ and the formula $\mathbb{E}[(\mathcal{N}(0,\sigma^2))^4] = 3\sigma^4$ \cite{CasellaBerger}. Thus,
\begingroup\makeatletter\def\f@size{9}\check@mathfonts
\begin{align}
\mathbb{V}_{n,\mathbf{W}} \left[ K_\Phi(\mathbf{X},\mathbf{Y}) \right] \leq \dfrac{1}{\nu^3} \mathbb{E}_{n} \left[ \dfrac{1}{\sigma^{8n}} \dfrac{\exp(-2\sigma^{-2})}{(\rho(n) n!)^2} 3^n \sigma^{4n} \left( \sum_{i,j=1}^d \log(\mathbf{X})^2_{ij} \log(\mathbf{Y})^2_{ij}\right)^n \right]. \label{eq:cactus3}
\end{align}\endgroup
Since $\sum_{i,j=1}^d \log(\mathbf{X})^2_{ij} \log(\mathbf{Y})^2_{ij} \leq \left( \sum_{i,j=1}^d \log(\mathbf{X})^2_{ij} \right) \left( \sum_{i,j=1}^d \log(\mathbf{Y})^2_{ij} \right) = 1$
due to the assumption of unitary Frobenius norm for both $ \log \mathbf{X}$ and $ \log \mathbf{Y}$, we get
\begin{align}
\mathbb{V}_{n,\mathbf{W}} \left[ K_\Phi(\mathbf{X},\mathbf{Y}) \right] \leq \dfrac{1}{\nu^3} \mathbb{E}_{n} \left[ \dfrac{1}{\sigma^{8n}} \dfrac{\exp(-2\sigma^{-2})}{(\rho(n) n!)^2} 3^n \sigma^{4n} \right]. \label{eq:cactus4}
\end{align}
We can now expand the expectation over $\rho$ in \eqref{eq:cactus4}, achieving
\begin{align}
\mathbb{V}_{n,\mathbf{W}} \left[ K_\Phi(\mathbf{X},\mathbf{Y}) \right] \leq \dfrac{\exp(-2\sigma^{-2})}{\nu^3} \sum_{n = 0}^\infty \left(\dfrac{3}{\sigma^4}\right)^n \dfrac{1}{n!} \sum_{n = 0}^\infty \dfrac{1}{\rho(n) n!}, \label{eq:cactus5}
\end{align}
since the series of the products is less than the product of the series, provided that both converge. This is actually true since, by exploiting the McLaurin expansion for the exponential function, we easily get $\sum_{n = 0}^\infty \left(\frac{3}{\sigma^4}\right)^n \frac{1}{n!} = \exp\left( \frac{3}{\sigma^4} \right)$. On the other hand, since $\rho$ is a probability distribution, it must be $
\lim_{n \to \infty } \frac{\rho(n+1)}{\rho(n)} = L$ where $0 < L \leq 1$, being $\mathbb{N}$ the support of $\rho$ and due to $\sum_{n = 0}^\infty \rho(n) = 1$. Then, since  $\lim_{n \to \infty} \frac{\rho(n)}{\rho(n+1)} = \frac{1}{L} < \infty$ and $\lim_{n \to \infty} \frac{1}{n +1} = 0$, by the ration criterion for positive-terms series \cite{Rudin}, there must exist a constant $\mathcal{C}_\rho > 0$ such that
\begin{equation}\label{eq:secondo}
\sum_{n = 0}^\infty \dfrac{1}{\rho(n) n!} = \mathcal{C}_\rho.
\end{equation}
Therefore, by combining \eqref{eq:secondo} in \eqref{eq:cactus5}, we obtain
\begin{align}
\mathbb{V}_{n,\mathbf{W}} \left[ K_\Phi(\mathbf{X},\mathbf{Y}) \right] \leq \dfrac{\exp(-2\sigma^{-2})}{\nu^3} \exp\left(\dfrac{3}{\sigma^4}\right) \mathcal{C}_\rho = \dfrac{\mathcal{C}_\rho}{\nu^3} \exp\left(\dfrac{3 - 2\sigma^2}{\sigma^4}\right), \nonumber
\end{align}
which  is the thesis.
\endproof

\hrulefill

\bibliographystyle{splncs03}
\bibliography{biblio}

\begin{thebibliography}{10}
\providecommand{\url}[1]{\texttt{#1}}
\providecommand{\urlprefix}{URL }

\bibitem{Bach:ICML05}
Bach, F.R., Jordan, M.I.: Predictive low-rank decomposition for kernel methods.
  In: ICML (2005)

\bibitem{Bishop}
Bishop, C.M.: Pattern Recognition and Machine Learning - Information Science
  and Statistics. Springer-Verlag New York, Inc. (2006)

\bibitem{G3D}
Bloom, V., Makris, D., Argyriou, V.: {G3D}: A gaming action dataset and real
  time action recognition evaluation framework. In: CVPR (2012)

\bibitem{CasellaBerger}
Casella, G., Berger, R.: Statistical Inference. Duxbury advanced series in
  statistics and decision sciences, Thomson Learning (2002)

\bibitem{Cavazza:ICPR16}
Cavazza, J., Zunino, A., San~Biagio, M., Murino, V.: Kernelized covariance for
  action recognition. In: ICPR (2016)

\bibitem{bellolui}
Cho, K., Chen, X.: Classifying and visualizing motion capture sequences using
  deep neural networks. CoRR  1306.3874 (2014)

\bibitem{liblinear}
Fan, R.E., Chang, K.W., Hsieh, C.J., Wang, X.R., Lin, C.J.: {LIBLINEAR}: A
  library for large linear classification. JMLR  9,  1871--1874 (2008)

\bibitem{MSRC}
Fothergill, S., Mentis, H.M., Kohli, P., Nowozin, S.: Instructing people for
  training gestural interactive systems. In: ACM-CHI (2012)

\bibitem{Harandi:CVPR14}
Harandi, M., Salzmann, M., Porikli, F.: Bregman divergences for infinite
  dimensional covariance matrices. In: CVPR (2014)

\bibitem{egizi}
Hussein, M., Torki, M., Gowayyed, M., El-Saban., M.: Human action recognition
  using a temporal hierarchy of covariance descriptors on 3d joint locations.
  IJCAI  (2013)

\bibitem{KK:AISTATS12}
Kar, P., Karnick, H.: Random feature maps for dot product kernels. In: AISTATS
  (2012)

\bibitem{ECCV16}
Koniusz, P., Cherian, A., Porikli, F.: Tensor representation via kernel
  linearization for action recognition from 3d skeletons. In: ECCV (2016)

\bibitem{Fastfood}
Le, Q., Sarlos, T., Smola, A.: Fastfood - approximating kernel expansion in
  loglinear time. In: ICML (2013)

\bibitem{Action3D}
Li, W., Zhang, Z., Liu, Z.: Action recognition based on a bag of 3d points. In:
  CVPR workshop (2010)

\bibitem{HDM-05}
M\"{u}ller, M., R\"{o}der, T., Clausen, M., Eberhardt, B., Kr\"{u}ger, B.,
  Weber, A.: {HDM-05} doc. In: Tech. Rep. (2007)

\bibitem{MSRPairs}
Oreifej, O., Liu., Z.: {HON4D}: Histogram of oriented {4D} normals for activity
  recognition from depth sequences. In: CVPR (2013)

\bibitem{RR:NIPS07}
Rahimi, A., Recth, B.: Random features for large-scale kernel machines. In:
  NIPS (2007)

\bibitem{Rudin}
Rudin, W.: Real and Complex Analysis, 3rd Ed. McGraw-Hill, Inc., New York, NY,
  USA (1987)

\bibitem{Florence3D}
Seidenari, L., Varano, V., Berretti, S., Bimbo, A.D., Pala, P.: Recognizing
  actions from depth cameras as weakly aligned multi-part bag-of-poses. In:
  CVPR workshops (2013)

\bibitem{Vemulapalli:CVPR14}
Vemulapalli, R., Arrate, F., Chellappa, R.: Human action recognition by
  representing 3d skeletons as points in a lie group. In: CVPR (June 2014)

\bibitem{survey}
Vrigkas, M., Nikou, C., Kakadiaris, I.A.: A review of human activity
  recognition methods. Front. robot. AI  2, ~28 (2015)

\bibitem{Wang:ICCV15}
Wang, L., Zhang, J., Zhou, L., Tang, C., Li, W.: Beyond covariance: Feature
  representation with nonlinear kernel matrices. In: ICCV (2015)

\bibitem{UTKinect}
Xia, L., Chen, C.C., Aggarwal, J.: View invariant human action recognition
  using histograms of {3D} joints. In: CVPR workshops (2012)

\bibitem{Zhang:ICML08}
Zhang, K., Tsang, I.W., Kwok, J.T.: Improved {Nystr\"{o}m} low-rank
  approximation. In: ICML (2008)

\end{thebibliography}

\end{document}